\documentclass[letterpaper, 10 pt, journal, twoside]{IEEEtran}

\usepackage{cite}
\usepackage{amsmath,amssymb,amsfonts}
\usepackage{algorithmic}
\usepackage{graphicx}
\usepackage{textcomp}
\usepackage{xcolor}
\usepackage{booktabs}
\usepackage{colortbl}
\usepackage{multirow}
\usepackage{orcidlink}
\hypersetup{hidelinks}

\usepackage[capitalize]{cleveref}
\crefname{section}{Sec.}{Secs.}
\Crefname{section}{Section}{Sections}
\Crefname{table}{Table}{Tables}
\crefname{table}{Tab.}{Tabs.}

\usepackage[export]{adjustbox}

\def\BibTeX{{\rm B\kern-.05em{\sc i\kern-.025em b}\kern-.08em
    T\kern-.1667em\lower.7ex\hbox{E}\kern-.125emX}}

\newcommand{\receiveddate}{April 27, 2026}
\newcommand{\reviseddate}{August 8, 2026}
\newcommand{\accepteddate}{September 9, 2026}
\newcommand{\acceptedmonthyear}{September, 2026}
\newcommand{\handlingeditor}{Ashis Banerjee}
\newcommand{\correspondingemail}{leandro.di.bella@vub.be}
\newcommand{\paperkeywords}{Intelligent Transportation Systems, Autonomous Vehicle Navigation, Deep Learning Methods}

\begin{document}

\title{FlowS: One-Step Motion Prediction via Local Transport Conditioning}

\author{Leandro Di Bella$^{1}$~\orcidlink{0009-0000-1731-7205},
    Adrian Munteanu$^{1}$~\orcidlink{0000-0001-7290-0428},~\IEEEmembership{Member,~IEEE},
    and Bruno Cornelis$^{1}$~\orcidlink{0000-0002-0688-8173}%
\thanks{Manuscript received: \receiveddate; Revised: \reviseddate; Accepted: \accepteddate.}%
\thanks{This paper was recommended for publication by Editor \handlingeditor\ upon evaluation of the Associate Editor and Reviewers' comments.}%
\thanks{$^{1}$Leandro Di Bella, Bruno Cornelis and Adrian Munteanu are with the Department of Electronics and Informatics, Vrije Universiteit Brussel, Pleinlaan 2, B-1050 Brussels, Belgium. Leandro Di Bella and Adrian Munteanu are also with IMEC, Kapeldreef 75, B-3001 Leuven, Belgium. Corresponding author: Leandro Di Bella (email: {\ttfamily\footnotesize\correspondingemail}).}%
\thanks{Digital Object Identifier (DOI): see top of this page.}%
}

\markboth{IEEE Robotics and Automation Letters. Preprint Version. Accepted \acceptedmonthyear}
{Di Bella \MakeLowercase{\textit{et al.}}: FlowS: One-Step Motion Prediction}

\maketitle

\begin{abstract}
Generative motion prediction must satisfy three simultaneous requirements for real-world autonomy: high accuracy, diverse multimodal futures, and strictly bounded latency. Diffusion models meet the first two but violate the third, requiring tens to hundreds of denoising steps. We identify a conditioning strategy that resolves this tension: \textit{single-step integration is accurate when the underlying transport problem is local}. A model that must both discover the correct behavioral mode and traverse a long displacement in one step accumulates large discretization errors; conditioning the base distribution to lie near plausible futures reduces the problem to short-range refinement, the regime where a single Euler step suffices. We instantiate this \emph{local transport conditioning} in FlowS, a conditional flow matching framework with two mechanisms. First, an online, scene-conditioned learned prior emits $K$ calibrated anchor trajectories per agent, each already near a plausible future, converting mode discovery into local correction. Second, a step-consistent displacement field enforces semigroup self-consistency, guaranteeing that a single step inherits multi-step accuracy. Crucially, anchoring this field at learned priors along straight-line paths yields a {stable, low-variance} training target, unlike prior self-consistency methods that suffer from {high-variance bootstrap} signals on curved diffusion paths. On the Waymo Open Motion Dataset, FlowS achieves state-of-the-art Soft mAP (0.4830) and mAP (0.4730) with a model ensemble at 75\,FPS with single-step inference, demonstrating that local transport conditioning makes one-step generative motion prediction practical for safety-critical autonomy. The code is publicly available at: https://github.com/leandro-svg/FlowS.git.
\end{abstract}

\begin{IEEEkeywords}
\paperkeywords
\end{IEEEkeywords}

\begin{figure*}
    \centering
    \includegraphics[width=0.78\linewidth]{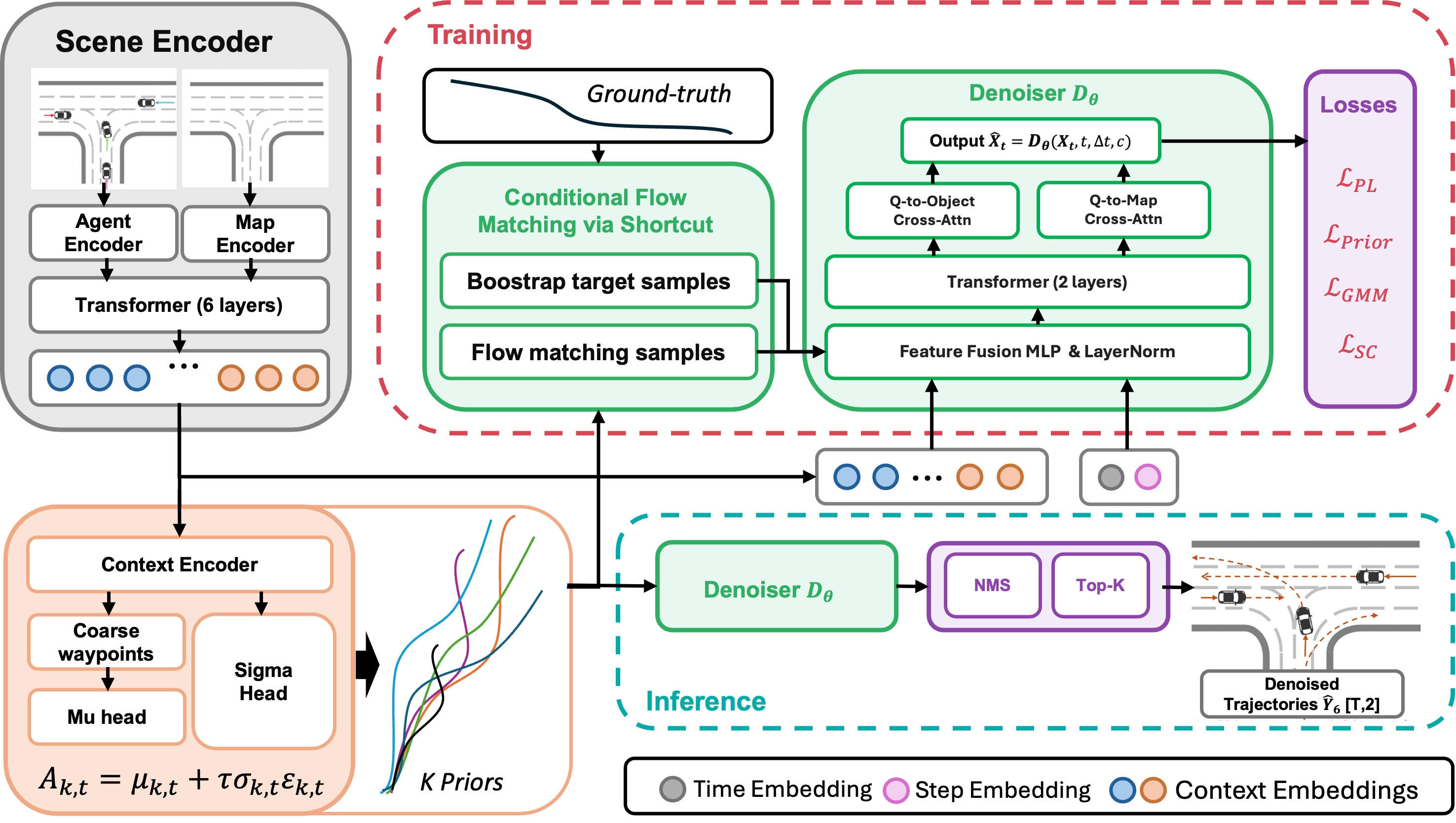}
    \caption{\textbf{FlowS architecture.} Scene inputs are processed by a Transformer-based encoder (Agent/Map). The \emph{Learned Prior} module replaces the standard Gaussian base by producing $K$ scene-conditioned anchor trajectories. During training, the decoder ($D_\theta$) learns flow dynamics from anchors to ground-truth distributions via a step-consistent displacement loss ($\mathcal{L}_{step}$). At inference, a single forward evaluation produces multi-agent trajectory predictions, filtered by NMS.}
    \label{fig:overview}
    \vspace{-0.5cm}
\end{figure*}

\section{Introduction}
\IEEEPARstart{M}{otion} prediction for autonomous driving must satisfy three simultaneous requirements: high accuracy, multimodal coverage, and strictly bounded latency. The essential challenge is clear: given the recent scene with spatial positions and movement dynamics, predict where all agents will be within the next few seconds. A forecast delivered after the next control update is functionally useless: if a pedestrian steps off the curb, the braking decision cannot wait. Meeting all three simultaneously remains challenging, and most methods prioritize some of these criteria at the expense of others. Recent progress has focused on generative models~\cite{ho2020denoising, lipman2022flow}, which excel at capturing the complex, multi-agent futures required in everyday driving. Among these, diffusion models stand out for their strong performance in both accuracy and diversity, as seen in frameworks like MotionDiffuser~\cite{jiang2023motiondiffuser, rowe2025scenario, mao2023leapfrog, gu2022stochastic}. Yet, this strength comes at a cost: diffusion forecasters generate realistic futures by iteratively refining samples across many denoising steps, often exceeding timing budgets imposed by industrial autonomy stacks.

Efforts to speed up sampling, such as DiffusionDrive~\cite{liao2025diffusiondrive} (truncated schedules from learned anchors) and BridgeDrive~\cite{liu2025bridgedrive} (diffusion bridges with matched PF-ODE sampling), have reduced steps but still require multiple refinement cycles and cannot guarantee single-cycle latency. This observation reveals a strategy that we believe is fundamental to few-step generative prediction: \emph{Single-step integration is accurate when the transport problem is conditioned to be local.}

Flow matching~\cite{lipman2022flow, liu2022flow, tong2023conditional} provides a more direct path: it learns velocity fields for deterministic trajectory transport without diffusion schedules. Recent works such as \mbox{TrajFlow}~\cite{yan2025trajflow} and GoalFlow~\cite{xing2025goalflow} demonstrate strong accuracy with fewer steps, yet both still degrade under aggressive step reduction. The common failure mode is clear: when initialized from a scene-agnostic Gaussian $Z\sim\mathcal{N}(0,I)$, the velocity field must simultaneously discover the correct behavioral mode and traverse a long displacement, a compound task that amplifies the discretization error in a single integration step. Conditioning the base distribution to lie near plausible futures reduces the transport to a short local correction, the regime in which single-step integration is accurate. This \emph{local transport conditioning strategy} is the conceptual core of our work. FlowS implements it with two complementary mechanisms:
\begin{itemize}
  \item \textbf{Scene-conditioned learned prior.} Rather than starting from $\mathcal{N}(0,I)$, we learn an online, scene-conditioned multimodal prior that emits $K$ calibrated anchor trajectories per agent, each already near a plausible future. This converts the global mode-discovery problem into local correction, satisfying the locality condition.
  
  \item \textbf{Step-consistent displacement field.} We train a semigroup-consistent transport field that enforces $T_{\Delta t} \circ T_{\Delta t} \approx T_{2\Delta t}$, so that a single step faithfully approximates multi-step dynamics. Anchoring this field at learned priors along straight-line paths produces a {stable, low-variance} self-consistency target throughout training, unlike prior formulations that bootstrap from curved diffusion paths and suffer from moving-target instability.
  
  \item \textbf{Real-time single-step prediction.} FlowS achieves state-of-the-art mAP on the Waymo Open Motion Dataset at $\sim$75\,FPS with single-step inference, demonstrating that local transport conditioning makes one-step generative motion prediction practical.
\end{itemize}

\section{Related Work}
Transformer-based deterministic forecasters dominate modern motion prediction: the Motion Transformers (MTR) family \cite{shi2022motion, shi2024mtr++} pioneered learnable intention queries decoupling goal localization from trajectory refinement, Wayformer \cite{nayakanti2022wayformer} unifies modality-agnostic attention, QCNet \cite{zhou2023query} introduces query-centric designs, and MotionLM \cite{seff2023motionlm} and SMART \cite{wu2024smart} reframe prediction as token-sequence modeling. Despite strong performance, these approaches output point estimates or fixed mode sets, limiting their capacity to capture multi-agent uncertainty.

Diffusion-based generative forecasting addresses this by modeling trajectory distributions directly. Building on DDPM \cite{ho2020denoising}, MotionDiffuser \cite{jiang2023motiondiffuser} produces permutation-invariant joint distributions via iterative refinement, while Leapfrog \cite{mao2023leapfrog} and \cite{gu2022stochastic} develop efficient stochastic variants. However, diffusion methods typically require 50--100 denoising steps, posing a prohibitive barrier for real-time autonomy.

Acceleration methods attempt to reduce this cost. DiffusionDrive \cite{liao2025diffusiondrive} initializes from pre-learnt multi-mode anchors with truncated schedules, cutting steps but introducing forward-reverse mismatches. BridgeDrive \cite{liu2025bridgedrive} formalizes this as a diffusion bridge with matched Probability Flow ODE sampling, restoring theoretical consistency. However, even \mbox{5--10} iterations cannot guarantee the real-time capability required by safety-critical systems.

Flow matching methods, TrajFlow~\cite{yan2025trajflow}, FlowDrive~\cite{wang2025flowdrive}, GoalFlow~\cite{xing2025goalflow}, learn direct velocity fields for deterministic transport. GoalFlow achieves strong accuracy in fewer steps, but initializing from Gaussian distributions limits single-step prediction: the field must locate and reach plausible modes in one pass.

Closest to our step-consistency mechanism are consistency models~\cite{song2023consistency}, which train a network to map any point of a probability-flow ODE directly to its endpoint, and Consistency Policy~\cite{prasad2024consistencypolicy}, which brought consistency distillation to real-time visuomotor control. FlowS differs in two ways: (i) consistency is enforced along \emph{anchored} straight-line transport paths whose conditional target velocity is constant, yielding a stable, low-variance self-consistency target rather than a distillation signal from a pretrained diffusion teacher; and (ii) the base distribution itself is learned and scene-conditioned, so the one-step map performs local correction rather than full transport from noise.

FlowS addresses the gap left by all of the above: the learned prior provides scene-adapted warm starts that shorten transport to the regime where single-step integration is accurate, while the semigroup-consistent displacement field absorbs marginal curvature, yielding, together with conditional flow matching~\cite{tong2023conditional, liu2022flow}, a unified framework for strictly bounded-latency multimodal forecasting.


\section{Method}\label{sec:method}
\subsection{Overview}
FlowS turns motion prediction into a short, one-step correction. A scene encoder summarizes each agent's surroundings; a learned prior proposes $K$ anchor trajectories that already lie near plausible futures; and a single displacement step nudges each anchor onto its predicted trajectory. Because the anchors start close to the target, one Euler step suffices, and a step-consistency constraint guarantees that this single step matches what a chain of many small steps would produce, yielding state-of-the-art accuracy at one network evaluation per prediction.
The pipeline (Fig.~\ref{fig:overview}) comprises three stages: (1) a symmetric scene encoder aggregating agent histories and HD-map polylines into per-agent context features $c\in\mathbb{R}^{D}$, which condition all subsequent modules (Sec.~\ref{subsec:backbone}); (2) a learned, scene-conditioned prior emitting $K$ calibrated anchor trajectories per agent (Sec.~\ref{subsec:learned_prior}); and (3) a step-consistent displacement field trained along straight-line paths that admits a true single-step limit (Sec.~\ref{subsec:shortcut}). A multi-shot decoder produces final candidate trajectories and calibrated scores (Sec.~\ref{subsec:backbone}).
 {\paragraph*{Notation.}
Three distinct quantities recur below.
(i)~\textbf{Flow model} $v_\phi(X_t,t,c)$: the network trained by Conditional Flow Matching (CFM) to estimate the expected velocity $\mathbb{E}[u_t \mid X_t, c]$~\cite{lipman2022flow,tong2023conditional,frans2024one}.
(ii)~\textbf{Velocity target} $u_t = Y_{\mathrm{GT}}-A_k$ (or $Y-Z$ from Gaussian noise): the fixed, time-independent supervision signal, the displacement from base sample to ground truth, constant along each Optimal Transport (OT) path.
(iii)~\textbf{Displacement field} $s_\psi(X_t,t,d,c)$: our finite-step predictor (Sec.~\ref{subsec:shortcut}) covering a fraction $d$ of the remaining path in one evaluation, with $\lim_{d\to 0}s_\psi = v_\phi$. The decoder $D_\theta$ (Fig.~\ref{fig:overview}) implements both fields with shared parameters, distinguished by the step-size token $d$; throughout, $c\!\in\!\mathbb{R}^D$ is the per-agent scene context from the encoder.}

\begin{figure}[t]
    \centering
    \includegraphics[width=\columnwidth]{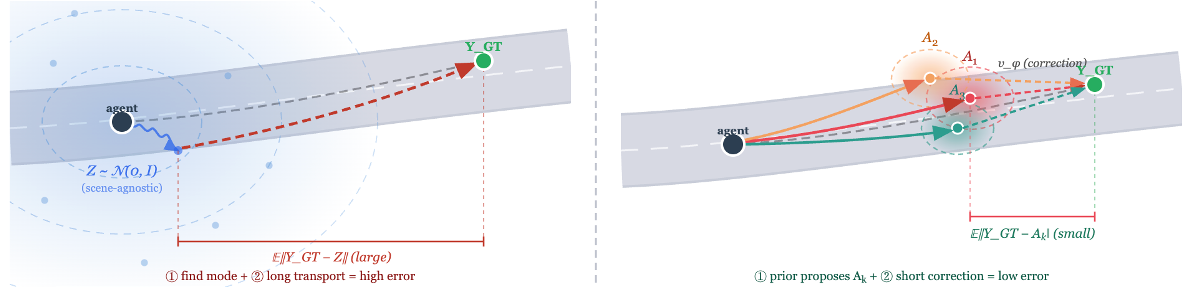}
    \caption{\textbf{Why a learned prior helps one-step flow matching.} \emph{Left:} a scene-agnostic sample $Z\sim\mathcal{N}(0,I)$ can lie far from the target, so a single update must both identify the correct mode and traverse a long displacement, increasing error. \emph{Right:} scene-conditioned anchors $A_k$ already lie near plausible futures, so one-step CFM only needs a short corrective transport: $\|Y_{\mathrm{GT}}-A_k\| \ll \|Y_{\mathrm{GT}}-Z\|$.}
    \label{fig:gaussianvsprior}
    \vspace{-0.5cm}
\end{figure}

\subsection{Preliminaries: Conditional Flow Matching}
\label{subsec:prelim_cfm}
We briefly review Conditional Flow Matching (CFM)~\cite{lipman2022flow, tong2023conditional, liu2022flow}, which provides the foundation for our contributions.
Let $Z\!\sim\!p_0$ be a sample from a base distribution (e.g.\ $p_0=\mathcal{N}(0,I)$), and let $Y\!\sim\!p_{\text{data}}$ be a ground-truth future trajectory in $\mathbb{R}^{T\times2}$ (with $T$ the number of prediction timesteps and 2 the spatial coordinates). The per-agent scene context $c\!\in\!\mathbb{R}^{D}$ is produced by the encoder (Sec.~\ref{subsec:backbone}) and conditions all learned fields. For each agent, a base sample $Z$ is drawn independently and paired with that agent's ground-truth future $Y$, forming a per-sample coupling $(Z,Y)$. CFM learns a time-indexed  {flow model} $v_\phi$ by matching it to a  {known velocity target} along a family of conditional probability paths $\{p_t(\cdot\,|\,Z,Y)\}_{t\in[0,1]}$ that smoothly interpolate between $p_0$ (at $t{=}0$) and $p_{\text{data}}$ (at $t{=}1$):
\vspace{-0.1cm}
\begin{equation}
\mathcal{L}_{\text{CFM}}
\;=\;
\mathbb{E}_{\substack{Z\sim p_0,\; Y\sim p_{\text{data}}\\ t\sim\mathcal{U}[0,1]}}
\Big\|\, v_\phi\big(X_t,t,c\big) - u_t\big(X_t\,|\,Z,Y\big)\,\Big\|_2^2 .
\label{eq:cfm_general_base}
\end{equation}
\vspace{-0.4cm}

A convenient choice is the optimal transport (OT) straight-line \cite{lipman2022flow, liu2022flow} path  between $Z$ and $Y$:
\vspace{-0.4cm}

\begin{equation}
X_t \;=\; (1-t)\,Z + t\,Y,
\qquad
u_t\big(X_t\,|\,Z,Y\big) \;=\; Y - Z,
\label{eq:ot_path_base}
\end{equation}
\vspace{-0.4cm}

which yields a \emph{time-independent} target velocity $u_t$ and closed-form supervision when substituted into~\eqref{eq:cfm_general_base}. In a \emph{one-step} regime, the flow is applied once from the base sample ($t{=}0$): $Y_{\text{pred}} = Z + v_\phi(Z,0,c)$.

\subsection{Learned Scene-Conditioned Prior}
\label{subsec:learned_prior}

 {Starting CFM from $Z\!\sim\!\mathcal{N}(0,I)$ forces a single update to both identify the correct mode and traverse a long displacement (Fig.~\ref{fig:gaussianvsprior}, left), amplifying truncation error. We instead learn a scene-conditioned multimodal prior that emits anchors $A_k$ lying near plausible futures (Fig.~\ref{fig:gaussianvsprior}, right), reducing one-step CFM to a short corrective transport:}

\vspace{-0.3cm}

\begin{equation}
\mathbb{E}\,\min_{k}\|Y_{\mathrm{GT}} - A_k\|_2 \;\ll\; \mathbb{E}\,\|Y_{\mathrm{GT}} - Z\|_2,
\label{eq:transport_reduction}
\end{equation}
\noindent where $\|\cdot\|_2$ denotes the $\ell_2$ norm over the full trajectory $\mathbb{R}^{T\times2}$ and the expectation is taken over agents and scenes.

\begin{figure}[t]
    \centering
    \includegraphics[width=\columnwidth,trim=40 40 0 160,clip]{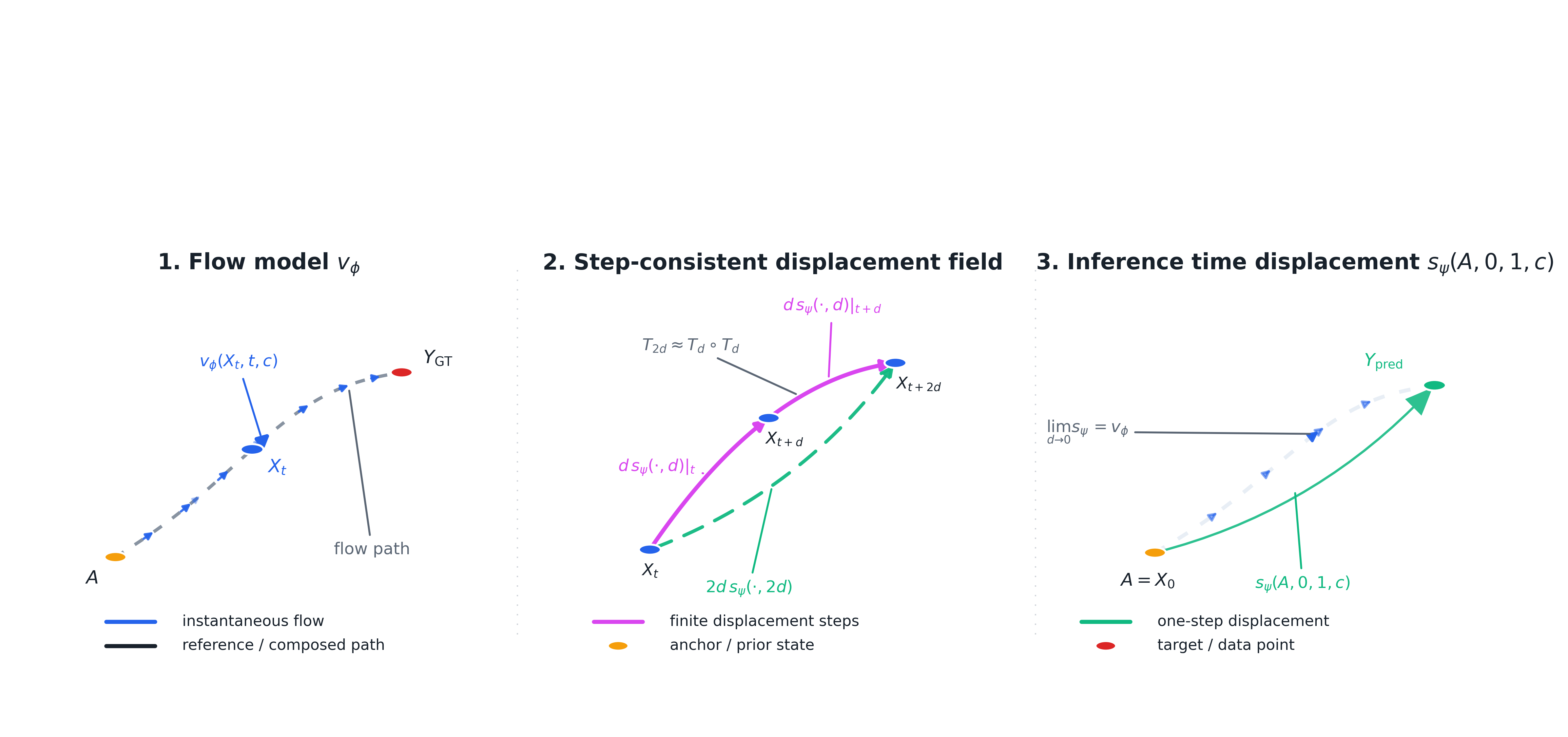}
    \caption{\textbf{Step-consistent displacement field.} \emph{Left:} the CFM velocity field $v_\phi$ along the path from anchor $A$ to the target. \emph{Middle:} semigroup consistency enforces that one $2d$ update matches two composed $d$ updates. \emph{Right:} at inference, $s_\psi(A,0,1,c)$ realizes the full transport in one pass while preserving $\lim_{d\to 0}s_\psi = v_\phi$.}
    \label{fig:shortcut_theory}
    \vspace{-0.5cm}
\end{figure}

Since the one-step Euler error scales directly with transport distance at $\Delta t\!=\!1$, this scene-dependent base converts the problem from global mode search into local refinement, the regime where single-step integration is accurate. Unlike offline anchor clustering \cite{liao2025diffusiondrive, jiang2025flowdrive}, we therefore learn an \emph{online scene-conditioned, multimodal prior} that emits $K$ calibrated hypotheses per agent and is directly compatible with one-step CFM. In practice $K\!=\!64$, i.e.\ one anchor per decoder query ($N_q\!=\!K$); the final prediction is reduced to the $6$ hypotheses required by the official WOMD protocol by non-maximum suppression, so the prior's anchor count and the number of submitted hypotheses are two distinct quantities.

Given scene features $c\in\mathbb{R}^{D}$, the prior emits a $K$-component mixture of mode-specialized \emph{anchors}. Each component is conditioned on $c$ and produces outputs that vary across trajectory timesteps $n\!\in\!\{1,\dots,T\}$: a mean trajectory, a per-timestep uncertainty, and a mode weight:

\vspace{-0.3cm}

\begin{equation}
\big\{\mu_k(c)\in\mathbb{R}^{T\times2},\;\sigma_k(c)\in{\mathbb{R}^{T\times2}_{\ge 0}}, \pi_k(c) \big\}_{k=1}^{K}.
\label{eq:prior_outputs}
\end{equation}
\vspace{-0.4cm}

\noindent Here $\mu_k(c) = [\mu_{k,1}(c),\dots,\mu_{k,T}(c)]$ and $\sigma_k(c) = [\sigma_{k,1}(c),\dots,\sigma_{k,T}(c)]$ are full-length trajectory sequences, where the subscript $n\!\in\!\{1,\dots,T\}$ indexes the \emph{prediction} timestep along the forecast horizon, not the flow time $t\!\in\![0,1]$ used in CFM. The network predicts each $\mu_k(c)$ and $\sigma_k(c)$ jointly in a single forward pass. Both are \emph{scene-conditioned} (functions of $c$) and \emph{per-timestep} (the network outputs a distinct value for each prediction step $n${, with separate scales for each spatial coordinate}): $\sigma_{k,n}$ is initialized to increase with $n$, reflecting that near-future positions are inherently more predictable than distant ones.

Mode specialization is implemented via $K$ learned queries $\{q_k\}$ with lightweight heads producing $\mu_k,\sigma_k$ (the FiLM parameterization and head architecture are detailed in Appendix~\ref{app:impl}). Equation~\eqref{eq:prior_anchor} is the only stochastic step. Sampling uses a single reparameterization, controlled by a temperature scalar $\tau\!>\!0$:
\vspace{-0.4cm}

\begin{equation}
A_k \;=\; \mu_k(c) + \big(\tau\,\sigma_k(c)\big)\odot \varepsilon_k, \text{    }\varepsilon_k\sim{\mathcal{N}(0,I_{2T})}.
\label{eq:prior_anchor}
\end{equation}
\vspace{-0.4cm}

\noindent In summary, the prior network defines the $K$-component mixture distribution; each anchor $A_k$ is a single sample drawn from mode $k$ of this distribution. The flow model $v_\phi$ then operates on each $A_k$.

\paragraph*{{Anchor-conditioned CFM in one-step}}
Conditioned on $A_k$, we train the flow model $v_\phi$ along the straight-line path to the ground truth:
\vspace{-0.4cm}

\begin{equation}
{X_t=(1-t)A_k+tY_{\text{GT}},\qquad u_t=Y_{\text{GT}}-A_k,}
\label{eq:anchor_path}
\end{equation}
 {where $u_t$ is the velocity target (the training signal for $v_\phi$, constant along this path).} This yields the one-step predictor:
\begin{equation}
Y_k \;=\; A_k + v_\phi(A_k,0,c).
\label{eq:onestep_from_anchor}
\end{equation}
\vspace{-0.4cm}

\noindent  {Starting from scene-adapted $A_k$ shortens transport, reducing one-step truncation error; $(\mu_k,\sigma_k)$ define mode geometry and dispersion while $v_\phi$ acts as a local refinement operator around each anchor.}

\paragraph{Prior training.} The prior is trained with four complementary objectives. The per-mode Gaussian NLL (element-wise along the trajectory):
\vspace{-0.3cm}

\begin{equation}
\mathrm{NLL}_k \;=\; \tfrac12 \left\|\frac{Y_{\text{GT}} - \mu_k}{\sigma_k}\right\|_2^2 + {\sum_{n=1}^{T}\sum_{j=1}^{2}\log\sigma_{k,n,j}}\,,
\label{eq:prior_nll}
\end{equation}

optimized as $\mathcal{L}_{\text{nll}} = \min_{k}\mathrm{NLL}_k$ to ensure at least one accurate hypothesis. A mixture calibration loss aligns predicted weights $\pi$ with soft targets $\hat{\pi}_k = \operatorname{softmax}(-\mathrm{NLL}_k/\tau_{\text{mix}})$ via KL divergence:

\vspace{-0.3cm}

\begin{equation}
\mathcal{L}_{\text{mix}} \;=\; D_{\mathrm{KL}}\bigl( \hat{\pi} \,\|\, \pi \bigr) \;=\; \sum\nolimits_{k=1}^{K}\hat{\pi}_k\log(\hat{\pi}_k/\pi_k).
\end{equation}
\vspace{-0.4cm}

The prior defines a $K$-component mixture distribution over anchor trajectories; at inference, rather than sampling a single component, we evaluate all $K$ modes in parallel; the mixture weights $\pi_k$ serve only as a training-time calibration target, while the confidence scores used for NMS selection are produced by the decoder's confidence head (Sec.~\ref{subsec:backbone}). A negative-entropy penalty $\mathcal{L}_{\text{ent}}$ prevents variance collapse, and a bounded hinge repulsion $\mathcal{L}_{\text{div}}$ enforces mode diversity. In plain terms, the prior is trained so that at least one anchor lands near the true future, its confidence weights are calibrated, and its modes stay diverse. The total prior objective is:
\vspace{-0.3cm}

\begin{equation}
{\mathcal{L}_{\text{prior}}
\;=\; \lambda_{\text{nll}}\mathcal{L}_{\text{nll}}
\;+\; \lambda_{\text{mix}}\mathcal{L}_{\text{mix}}
\;+\; \lambda_{\text{ent}}\mathcal{L}_{\text{ent}}
\;+\; \lambda_{\text{div}}\mathcal{L}_{\text{div}}}
\label{eq:prior_total}
\end{equation}
\vspace{-0.6cm}

\subsection{Step-Consistent Displacement Field}
\label{subsec:shortcut}

 {Even with straight conditional paths $(1{-}t)A_k + tY_{\mathrm{GT}}$, the \emph{marginal} velocity field $v_\phi$ averages over conditional paths that cross in state space, inducing effective curvature and one-step discretization error or averaging bias in one-step prediction. The prior mitigates this by spatially separating paths across modes, but residual curvature remains where anchors share intermediate states. We therefore introduce $s_\psi$, a step-conditioned displacement field that enforces compositional self-consistency yielding a field that implicitly absorbs the cumulative effect of marginal curvature ~\cite{yang2024consistencyfm}: two consecutive small steps must agree with one large step, so the $d{=}1$ output reproduces the endpoint a chain of locally accurate steps would reach. Concretely, $s_\psi(X_t, t, d, c)$ predicts the displacement over a fraction $d\!\in\![0,1]$ of the remaining path, and is directly compatible with one-step inference at $d{=}1$.}
Given a prior anchor $A$ (Sec.~\ref{subsec:learned_prior}), the encoded context $c$, flow time $t\!\in\![0,1]$, and a step-size token $d\!\in\!{[0,1]}$, we predict the $d$-ahead displacement with a step-conditioned field:
\begin{equation}
X_{t+d} \;=\; X_t \;+\; d\,s_\psi(X_t, t, d, c), \\
\end{equation}
\begin{equation}
\text{with }\lim_{d\to 0} s_\psi(X_t, t, d, c) \;=\; v_\phi(X_t, t, c).
\label{eq:shortcut_update}
\end{equation}

where $v_\phi$ is the flow model.  {At $d{=}0$, $s_\psi$ recovers the flow model output; at $d{>}0$ it predicts a scale-aware finite displacement. Time $t$ and step size $d$ are embedded via sinusoidal encodings and concatenated to the conditioning. We impose compositional self-consistency: two consecutive $d$-sized updates should match a single $2d$ update, i.e.\ $T_{\Delta t}\!\circ\!T_{\Delta t}\!\approx\!T_{2\Delta t}$ (Fig.~\ref{fig:shortcut_theory}).}

During training, we blend flow-step CFM supervision with large-step self-consistency. Sampling $t\!\sim\!\mathcal{U}[0,1-d]$, $d$ from a small set, and a prior anchor $A_k$:

\begin{align}
\mathcal{L}_{\text{flow}} 
&= \mathbb{E}\,\bigl\|\, s_\psi(X_t, t, 0, c) - (Y_{\text{GT}} - A_k)\,\bigr\|_2^2,
\label{eq:shortcut_update_s1}
\\[-2pt]
\mathcal{L}_{\text{cons}}
&= \mathbb{E}\,\bigl\|\, s_\psi(X_t, t, 2d, c) - s_{\text{target}}(X_t, t, d, c)\,\bigr\|_2^2
\label{eq:shortcut_update_s2}
\\[-2pt]
\text{with } s_{\text{target}} &= \tfrac12\,s_\psi(X_t, t, d, c) \notag \\
&\quad + \tfrac12\,s_\psi\!\Bigl(X_t + {d}\,s_\psi(X_t, t, d, c),\, t+{d},\, d,\, c\Bigr),
\label{eq:shortcut_update_s3}
\\[-2pt]
\mathcal{L}_{\text{step}}
&= {\lambda_{\text{flow}}\,\mathcal{L}_{\text{flow}}+\lambda_{\text{cons}}\,\mathcal{L}_{\text{cons}}.}
\label{eq:shortcut_update_s4}
\end{align}

Here $\mathcal{L}_{\text{flow}}$ anchors $s_\psi(\cdot,\cdot,0,\cdot)$ to the closed-form straight-line target (the same target used by $v_\phi$), ensuring $\lim_{d\!\to\!0}s_\psi = v_\phi$; $\mathcal{L}_{\text{cons}}$ enforces large-step coherence via~\eqref{eq:shortcut_update_s2}.  The self-consistency target $s_{\text{target}}$ in~\eqref{eq:shortcut_update_s3} is computed from an exponential moving average (EMA) of $s_\psi$, stabilizing the bootstrap signal during early training when the displacement field has not yet converged.
\vspace{-0.1cm}

 {All $s_\psi$ updates are anchored at prior anchors $A_k$, not Gaussian noise, with $X_t$ sampled along $(1{-}t)A_k + t\,Y_{\text{GT}}$. Training samples $t\!\sim\!\mathcal{U}[0,1{-}d]$ to expose the field to intermediate states; the EMA teacher is evaluated on the same anchor-rooted states, so both teacher and student see an identical input distribution. At inference we set $d{=}1$:}
\begin{equation}
{Y_{\text{pred},k} \;=\; A_k + s_\psi(A_k, 0, 1, c).}
\label{eq:shortcut_infer}
\end{equation}
 {The displacement field thus matches CFM in the infinitesimal limit, composes coherently for long jumps via self-consistency, and enables anchored one-pass prediction. Formally, $T_{\Delta t}\!\circ\!T_{\Delta t}\!\approx\!T_{2\Delta t}$ is a discrete semigroup-consistency condition that generalises $v_\phi$ into a one-step integrator preserving the multi-step flow structure.}

\begin{table*}[t]
\centering
\setlength{\tabcolsep}{8pt}
\renewcommand{\arraystretch}{1.15}
\caption{\textbf{WOMD test leaderboard results.} Best in \textbf{bold}, second \underline{underlined}, third marked $^*$. FlowS achieves the highest mAP and Soft mAP with single-step inference ($\sim$75\,FPS).}
\label{tab:main_test}
\resizebox{\linewidth}{!}{%
\begin{tabular}{@{}llcccccc@{}}
\toprule
\textbf{Method} & \textbf{Venue} & \textbf{Soft mAP $\uparrow$} & \textbf{mAP $\uparrow$} & \textbf{minADE $\downarrow$} & \textbf{minFDE $\downarrow$} & \textbf{Miss Rate $\downarrow$} & \textbf{Overlap Rate $\downarrow$} \\
\midrule
RMP-YOLO (Ens.)~\cite{sun2025rmp}       & ICRA'25    & 0.4737$^*$ & 0.4531 & 0.5564$^*$ & \underline{1.1188} & \textbf{0.1084} & 0.1259 \\
ModeSeq (Ens.)~\cite{zhou2025modeseq}   & CVPR'25    & 0.4737$^*$ & \underline{0.4665} & 0.5680 & 1.1766 & 0.1204 & 0.1275 \\
BeTop~\cite{liu2024reasoning}            & NeurIPS'24 & 0.4698 & 0.4587 & 0.5716 & 1.1668 & 0.1183 & 0.1272 \\
MGTR~\cite{gan2024multi}                 & ICRA'24    & 0.4599 & 0.4505 & 0.5918 & 1.2135 & 0.1298 & 0.1275 \\
EDA~\cite{lin2024eda}                    & AAAI'24    & 0.4596 & 0.4487 & 0.5718 & 1.1702 & 0.1169 & 0.1266 \\
MTR++~\cite{shi2024mtr++}                & TPAMI'24   & 0.4410 & 0.4329 & 0.5906 & 1.1939 & 0.1298 & 0.1281 \\
MTR~\cite{shi2022motion}                 & NeurIPS'22 & 0.4403 & 0.4249 & 0.5964 & 1.2039 & 0.1312 & 0.1274 \\
IMPACT (Ens.)~\cite{sun2025impact}       & arXiv'25   & \underline{0.4801} & 0.4598 & \underline{0.5563} & 1.1295$^*$ & \underline{0.1087} & 0.1258$^*$ \\
IMPACT~\cite{sun2025impact}              & IROS'25   & 0.4721 & 0.4609$^*$ & 0.5641 & 1.1540 & 0.1143 & \textbf{0.1255} \\
TrajFlow~\cite{yan2025trajflow}  & IROS'25   & 0.4572 & 0.4466 & 0.5811 & 1.1846 & 0.1193 & 0.1296 \\
TrajFlow (Ens.)~\cite{yan2025trajflow}  & IROS'25   & 0.4710 & 0.4604 & 0.5714 & 1.1667 & 0.1162 & 0.1272 \\
\midrule
\rowcolor{gray!15} \textbf{FlowS (Ours)} & - & 0.4658 & 0.4512 & 0.5756 & 1.1715 & 0.1182 & 0.1287 \\
\rowcolor{gray!15} \textbf{FlowS (Ours, Ens.)} & - & \textbf{0.4830} & \textbf{0.4730} & \textbf{0.5521} & \textbf{1.1183} & 0.1110$^*$ & \underline{0.1256} \\
\bottomrule
\end{tabular}
}
    \vspace{-0.5cm}\end{table*}

\paragraph*{Why anchored straight-line paths matter}
The straight-line transport $X_t = (1-t)A_k + t\,Y_{\mathrm{GT}}$ keeps the conditional target velocity $u_t = Y_{\mathrm{GT}} - A_k$ constant along each path~\cite{lipman2022flow, liu2022flow}, giving the consistency loss a fixed, closed-form target at every $t$: the EMA bootstrap signal tracks a well-defined quantity rather than the moving conditional target reported for self-consistency on curved diffusion paths~\cite{frans2024one}. On this foundation, the two components suppress \emph{distinct} single-step error sources: the prior reduces the transport \emph{distance} $\|Y_{\mathrm{GT}} - A_k\|$, which directly bounds the $d{=}1$ Euler residual regardless of field geometry, while $s_\psi$ absorbs the \emph{marginal curvature} induced by crossing paths via semigroup consistency, in the spirit of Consistency-FM~\cite{yang2024consistencyfm}. The $K$ mode-specialised anchors additionally separate conditional paths by behavioural mode, compressing the range over which $s_\psi$ must remain consistent; distance suppression and curvature absorption are therefore complementary.

\subsection{Backbone and Decoder}
\label{subsec:backbone}

For the encoder of FlowS, we adopt the symmetric MTR++ encoder \cite{shi2024mtr++}. Agent histories and HD-map polylines are vectorized into \emph{polyline tokens} and processed with query-centric self-attention in local coordinates (agent pose or lane tangent) using relative position/heading cues. This preserves locality, scales to large neighborhoods via local attention, and yields a shared scene representation
$
C \in \mathbb{R}^{N\times D},
$
from which we extract per-agent context features \(c \in \mathbb{R}^{D}\). These features are the sole inputs to our learned prior and one-step flow modules (cf.\ Fig.~\ref{fig:overview}).

For candidate generation, we follow the well-established \emph{multi-shot} decoder \cite{shi2022motion, shi2024mtr++, yan2025trajflow}. For each agent, we instantiate \(N_q\) learnable query tokens \(\{Q_i\}_{i=1}^{N_q}\) and apply stacked self-/cross-attention over \(C\). Three lightweight heads F produce candidate trajectories and scores:
\begin{equation}
\hat Y_k = \mathrm{F}_{\text{tr}}(Q),\;
S_k = \mathrm{F}_{\text{conf}}(Q),\;
r_k = \mathrm{F}_{\text{rank}}(Q).
\label{eq:decoder_heads}
\end{equation}
with $k=1,\dots,N_q$. Non-maximum suppression prunes redundant proposals to a fixed-size, diverse set. The confidence head $S_k$ (sigmoid) produces the per-hypothesis scores that drive NMS pruning; the rank head $r_k$ is additionally supervised by a Plackett--Luce (PL) ranking loss during training as a calibration regularizer~\cite{yan2025trajflow}, and is not used to re-rank at inference.

\begin{table}[t]
\centering
\setlength{\tabcolsep}{14pt}
\renewcommand{\arraystretch}{1.15}
\caption{\textbf{Per-class performance on the validation set of Waymo Open Motion Dataset.}}
\label{tab:womd_val}
\resizebox{\columnwidth}{!}{%

\begin{tabular}{@{}llccc@{}}
\toprule
\textbf{Setting} & \textbf{Category} & \textbf{mAP $\uparrow$} & \textbf{minADE $\downarrow$}  \\
\midrule
\multirow{4}{*}{\textbf{FlowS (Single)}}
  & Vehicle    & 0.4756 & 0.6707   \\
  & Pedestrian & 0.4808 & 0.3551  \\
  & Cyclist    & 0.4030 & 0.6969  \\
  & \textbf{Avg} & 0.4518 & 0.5745   \\
\midrule
\multirow{4}{*}{\textbf{FlowS (Ens.)}}
  & Vehicle    & 0.4963 & 0.6614   \\
  & Pedestrian & 0.5015 & 0.3346  \\
  & Cyclist    & 0.4152 & 0.6668  \\
  & \textbf{Avg} & \textbf{0.4710} & \textbf{0.5543}   \\

\bottomrule
\end{tabular}}
    \vspace{-0.2cm}

\end{table}

\begin{figure}[t]
    \centering
    \begin{minipage}[t]{0.44\columnwidth}
        \centering
        \includegraphics[width=\linewidth]{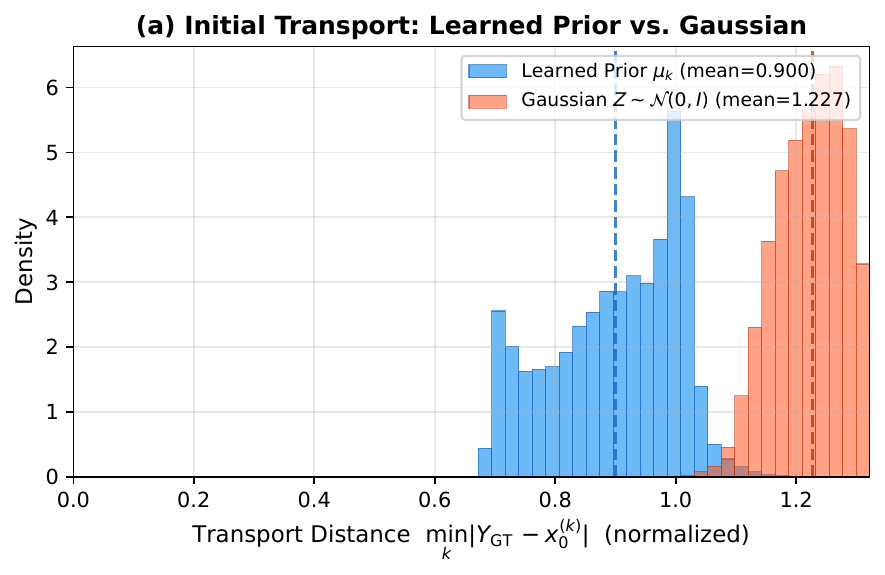}
        
    \end{minipage}\hfill
    \begin{minipage}[t]{0.44\columnwidth}
        \centering
        \includegraphics[width=\linewidth]{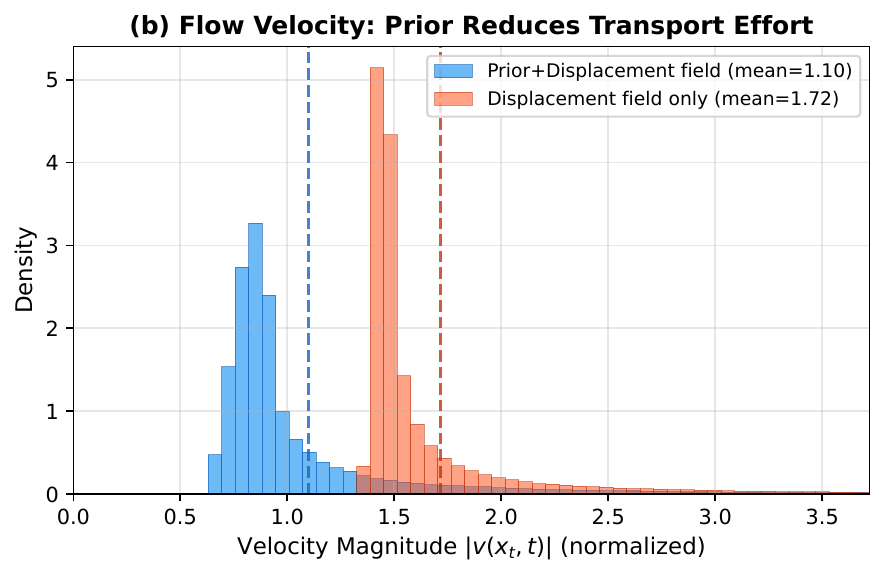}
    \end{minipage}
    \vspace{-0.3cm}
    
    \caption{\textbf{Prior--displacement field synergy.}
    \emph{Left:} Minimum transport distance to ground truth for the learned prior (blue, mean 0.78) vs.\ Gaussian $\mathcal{N}(0,I)$ (red, mean 1.23), showing a 36\% reduction that enables accurate one-step integration.
    \emph{Right:} One-step velocity magnitude for prior$+$displacement field (blue, mean 1.10) vs.\ displacement field only from $\mathcal{N}(0,I)$ (red, mean 1.72). The prior converts long-range transport into compact local refinement, the regime where step self-consistency is most effective.}
    \label{fig:synergy_evidence}
    \vspace{-0.5cm}
    
\end{figure}

\subsection{Implementation Details}
\label{subsec:dataset_metrics}

\textbf{Dataset: }We evaluate on the Waymo Open Motion Dataset (WOMD) \cite{ettinger2021large} motion prediction challenge: 9-second scenarios (1\,s history, 8\,s future) sampled at 10\,Hz, with official splits of 487k/44k/44k train/validation/test scenarios. Each scene requires forecasting up to 8 target agents over the full 8-second horizon.

\textbf{Metrics.} We report the standard WOMD metrics: minADE and minFDE (minimum displacement over $K$ hypotheses, agent-type averaged), Miss Rate, Overlap Rate, and the official (soft) mAP evaluated at 3, 5, and 8\,s across agent classes.

\textbf{Training.} We optimize with AdamW (weight decay $0.01$) at a batch size of $20$ per GPU on four A100 (80\,GB) GPUs (global batch $80$) for $50$ epochs, with a base learning rate of $1.5{\times}10^{-4}$ halved at epochs $\{20,25,30,35,48\}$ and gradient-norm clipping at $1.0$. Final predictions are selected by greedy NMS with an adaptive endpoint threshold in $[2.5,\,3.5]$\,m, retaining $6$ hypotheses per agent. Remaining default hyperparameters are listed in Appendix~\ref{app:impl}.

\begin{figure}[t]
    \centering
    \includegraphics[width=0.85\columnwidth,trim=0 5 0 40,clip]{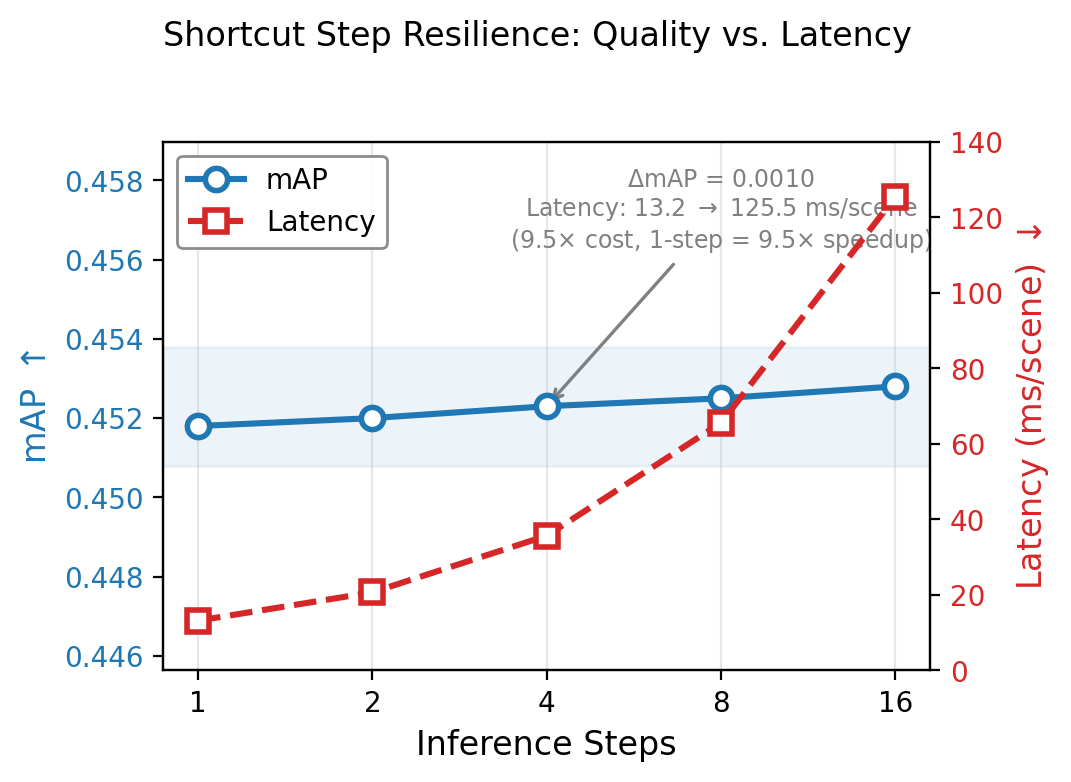}
    
    \caption{\textbf{Quality--latency trade-off.} mAP (blue, left axis) vs.\ latency (red, right axis) across 1--16 inference steps. The semigroup constraint distills multi-step accuracy into a single step: $\Delta$mAP$\,{=}\,0.0010$ with $9.5\times$ speedup.}
    \label{fig:shortcut_steps_tradeoff}
    \vspace{-0.5cm}
\end{figure}
\section{Experiments}

\subsection{Main Results}
We benchmark FlowS on WOMD (Tab.~\ref{tab:main_test} and Tab.~\ref{tab:womd_val}). On the official test leaderboard, FlowS (Ens.) achieves the highest Soft mAP (0.4830) and mAP (0.4730), surpassing all published entries including ensemble methods, while maintaining competitive displacement and miss rate. The single-model variant already reaches mAP 0.4512. 
    To assess training stability, we report validation metrics aggregated over $N\!=\!6$ independently trained runs, all using the prior$+$displacement field configuration: mAP $= 0.4518 \pm 0.0017$, minADE $= 0.5751 \pm 0.0065$, minFDE $= 1.1702 \pm 0.0152$, and Miss Rate $= 0.1180 \pm 0.0024$. The low standard deviations ($<\!0.4\%$ relative for mAP) indicate reliable convergence.

\begin{figure}[t]
    \centering
    \includegraphics[width=0.9\columnwidth,trim=0 169 0 0,clip]{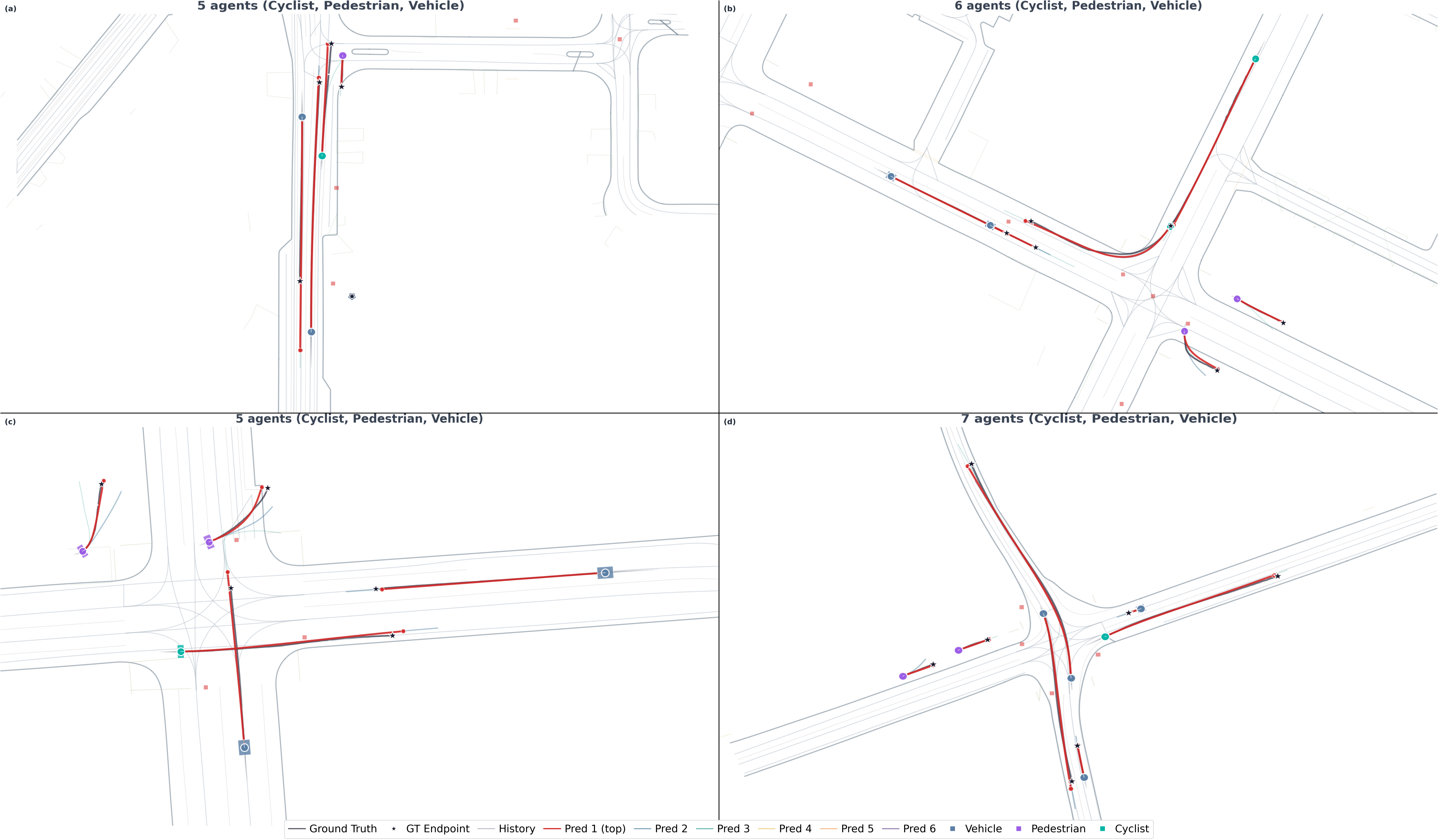}\\[2pt]
    \includegraphics[width=\columnwidth]{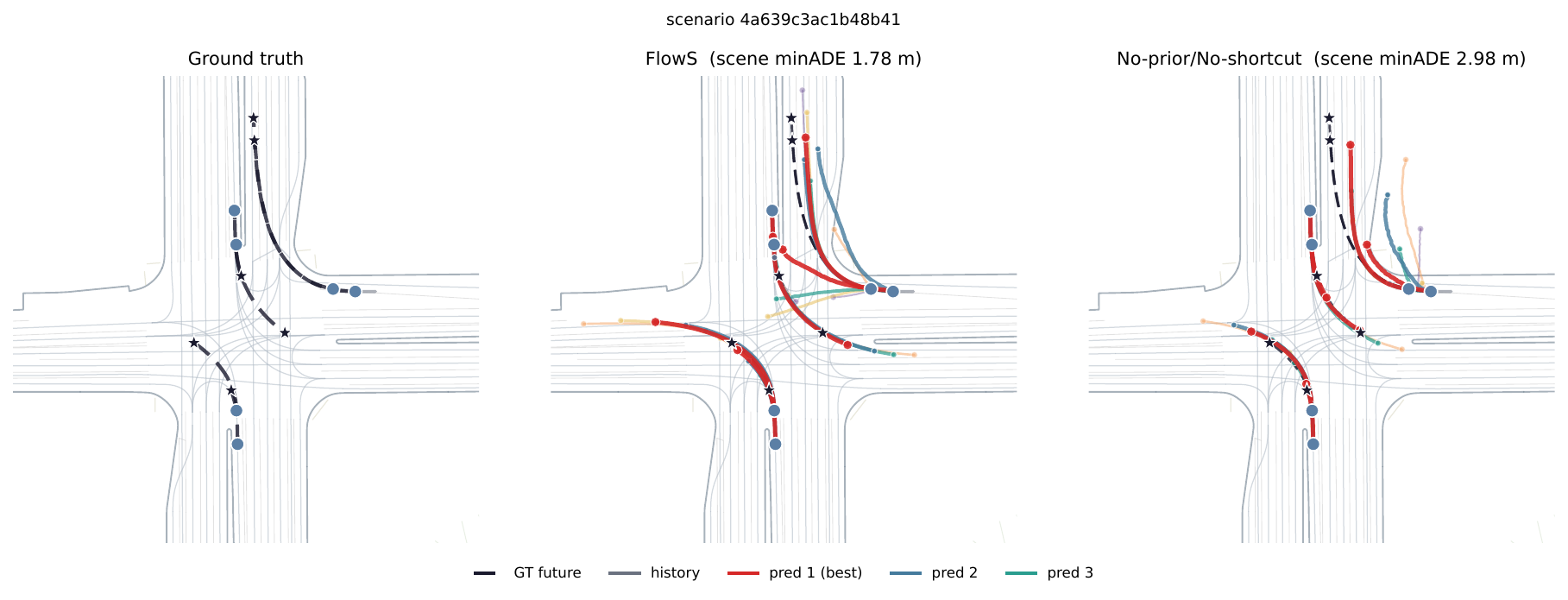}
    \vspace{-0.1cm}
    \caption{\textbf{Qualitative results on WOMD.} \emph{Top:} dense intersections, yielding maneuvers, and branching scenarios; despite single-step generation, trajectories remain smooth, lane-compliant, and multimodal. \emph{Bottom:} GT (left) $\mid$ FlowS (middle) $\mid$ no-prior/no-shortcut baseline (right) on a dense intersection; the $6$ predicted modes per agent are colored, the GT future dashed. FlowS: scene minADE $1.78$\,m; baseline: $2.98$\,m.}
    \label{fig:qualitative}
    \vspace{-0.5cm}
\end{figure}

\paragraph{Latency.} FlowS runs at $\sim$75\,FPS ($\sim$13.2\,ms/scene) on a single A100 with single-step inference, evaluated on the full WOMD validation set (44k scenes in 584\,s). Using the same measurement protocol as TrajFlow~\cite{yan2025trajflow}: MTR runs at 19.96\,ms/scene, TrajFlow at 16.44\,ms/scene, and FlowS at 13.25\,ms/scene, a 34\% reduction over MTR. The ensemble aggregates predictions from 6 independently trained models with varied hyperparameters (dropout rates, hidden dimensions, {flow settings}).

\subsection{Ablation Studies}

\begin{table}[t]
\centering
\setlength{\tabcolsep}{10pt}
\renewcommand{\arraystretch}{1.15}
\caption{\textbf{Component ablation} (single-step inference, validation set). Both components contribute complementary gains.}

\label{tab:waymo_best_map_ade}
\begin{tabular}{@{}llcc@{}}
\toprule
\textbf{Prior} & \textbf{Step-Consistent Field} & \textbf{mAP $\uparrow$} & \textbf{minADE $\downarrow$} \\
\midrule
$\times$ & $\times$ & 0.4376 & 0.5916 \\
$\times$ & $\checkmark$ & 0.4432 & 0.5850 \\
$\checkmark$ & $\times$ & {0.4458} & 0.5823 \\
$\checkmark$ & $\checkmark$ & \textbf{0.4518} & \textbf{0.5745} \\
\bottomrule
\end{tabular}
    \vspace{-0.2cm}

\end{table}
\paragraph{Component Contribution.} Tab.~\ref{tab:waymo_best_map_ade} ablates the two core components under single-step inference. Without either, baseline Gaussian flow matching achieves 0.4376 mAP. Adding the prior alone improves mAP by $+$0.0082 (shorter transport distance enables better mode capture). Adding the step-consistent field alone yields $+$0.0056 (compositional consistency corrects residual marginal curvature from path crossing). Together, they achieve 0.4518 mAP, the gains are complementary, not redundant.

\paragraph{Anchor-count ($K$) ablation.}
The prior emits $K$ anchors per agent ($K\!=\!N_q$), which NMS prunes to the $6$ scored hypotheses. Tab.~\ref{tab:k_ablation} ablates $K$ in two ways. \emph{(i) Inference-time budget:} keeping only the top-$K'$ anchors of the trained $K{=}64$ model degrades accuracy, and cyclists, the rarest, most multimodal class, lose coverage first (cyclist mAP $0.40\!\to\!0.06$ as $K'\!:64\!\to\!3$), directly supporting our limitation analysis; peak memory spans $5.2$--$10.2$\,GB, quantifying the coverage--cost trade-off. \emph{(ii) Retrained:} models trained from scratch with $K{=}16,32$ reach $0.4322$ and $0.4431$ mAP, a clean monotonic trend, with $K{=}32$ within $0.009$ of the full $K{=}64$ model ($0.4518$), so halving the anchor budget is nearly free when the modes are \emph{learned}, whereas naively subsampling to $32$ anchors at inference costs far more ($0.33$).

\begin{table}[t]
\centering

\setlength{\tabcolsep}{4pt}
\renewcommand{\arraystretch}{1.15}
\caption{\textbf{Anchor-count ($K$) ablation} (single-step, validation set). \emph{Top:} inference-time anchor budget $K'$ on the trained $K{=}64$ model (top-$K'$ modes kept before NMS; $K'{=}64$ is the full model). \emph{Bottom:} models \emph{retrained} from scratch with $K{=}16$ and $K{=}32$}
\label{tab:k_ablation}
\begin{tabular}{@{}lccccc@{}}
\toprule
$K'$ & mAP $\uparrow$ & Cyc.\ mAP $\uparrow$ & minADE $\downarrow$ & Thrpt.\ (s/s) $\uparrow$ & Mem.\ (GB) $\downarrow$ \\
\midrule
3  & 0.1114 & 0.0569 & 1.930 & 535 & 5.2 \\
6  & 0.1233 & 0.0685 & 1.504 & 521 & 5.3 \\
16 & 0.2368 & 0.1705 & 0.937 & 445 & 5.3 \\
32 & 0.3304 & 0.2560 & 0.727 & 381 & 6.1 \\
48 & 0.4130 & 0.3836 & 0.625 & 333 & 8.1 \\
64 & \textbf{0.4518} & \textbf{0.4030} & \textbf{0.574} & 288 & 10.2 \\
\midrule
\multicolumn{6}{@{}l}{\emph{Retrained} from scratch (vs.\ $K{=}64$ model, $0.4518$ mAP):} \\
$16^{\dagger\ddagger}$ & 0.4322 & -- & 0.614 & -- & -- \\
$32^{\dagger}$ & 0.4431 & -- & 0.589 & -- & -- \\
\bottomrule
\end{tabular}

\vspace{-0.4cm}
\end{table}

\paragraph{Prior--Displacement Field Synergy.}
Fig.~\ref{fig:synergy_evidence} explains why neither component alone suffices. The prior reduces mean transport distance by 36\% (0.78 vs.\ 1.23), but a single Euler step still accumulates discretization error without scale-aware correction. Conversely, the displacement field alone must bridge the full Gaussian-to-data gap (mean velocity 1.72), making its self-consistency target noisy. Together, the prior eliminates the \emph{mode-discovery} burden while the displacement field eliminates \emph{discretization error} on the resulting short transport (mean velocity 1.10). The prior thus serves double duty: it improves initialization \emph{and} stabilizes displacement field training.

\paragraph{Step Resilience.}
Fig.~\ref{fig:shortcut_steps_tradeoff} confirms that the semigroup constraint successfully distills multi-step accuracy into a single step. Sweeping inference steps from 1 to 16, mAP varies by only $\Delta$mAP$\,{=}\,0.0010$ while latency grows linearly from 3 to 28.7\,ms/agent, a $9.5{\times}$ speedup with negligible accuracy cost.

\paragraph{Drift by Scene Density.}
Grouping validation scenes by agent count, the learned prior improves minADE consistently across all density regimes, sparse (1--3 agents): $0.943$ vs.\ $0.983$ for the Gaussian base; medium (4--5): $0.843$ vs.\ $0.894$; dense (6+, 190k scenes): $0.734$ vs.\ $0.756$, a ${\sim}2.99\%$ relative gain that is largest in dense scenes where mode ambiguity is highest, with minFDE improving by $3.96\%$ in the dense regime ($1.017$ vs.\ $1.059$).

A per-case comparison against a $-$prior ablation (identical recipe, Gaussian base) confirms the prior's benefit is \textbf{non-uniform}: it is largest on rare high-curvature maneuvers (cyclist hard-turns improve minADE by $0.231$\,m, $-9.6\%$) and negligible on straight-line motion ($+0.002$\,m), consistent with the cyclist tail-coverage limitation discussed below.

\paragraph{Scale-Awareness and Per-Horizon Error.}
The displacement field is scale-aware: at the full step ($d\!=\!1$, $t\!=\!0$) its output matches the target in both magnitude and direction ($\cos\!=\!0.96$), with a stable $\ell_2$-to-target ($\approx\!0.16$) across all $t$, confirming a consistent endpoint prediction wherever it is queried. Per-horizon error remains low, median $1.52$\,m and cumulative ADE $0.58$\,m at the full $8$\,s horizon, so single-step generation does not disproportionately degrade long-horizon accuracy.

\paragraph{Qualitative Results}
Fig.~\ref{fig:qualitative} shows FlowS predictions across dense urban intersections, yielding maneuvers, and branching multi-agent scenarios. Despite single-step generation, the forecasted trajectories remain smooth, lane-compliant, and accurately capture multimodal branching behavior. On a dense intersection (Fig.~\ref{fig:qualitative}, bottom), FlowS stays lane-compliant and covers the multi-agent turning futures (scene minADE $1.78$\,m) where the no-prior/no-shortcut baseline (the $\times/\times$ ablation of Tab.~\ref{tab:waymo_best_map_ade}) scatters and misses tighter turns ($2.98$\,m).

\begin{figure}
    \centering
    \includegraphics[width=0.85\linewidth,trim={2 2cm {17cm} 8.3cm},clip]{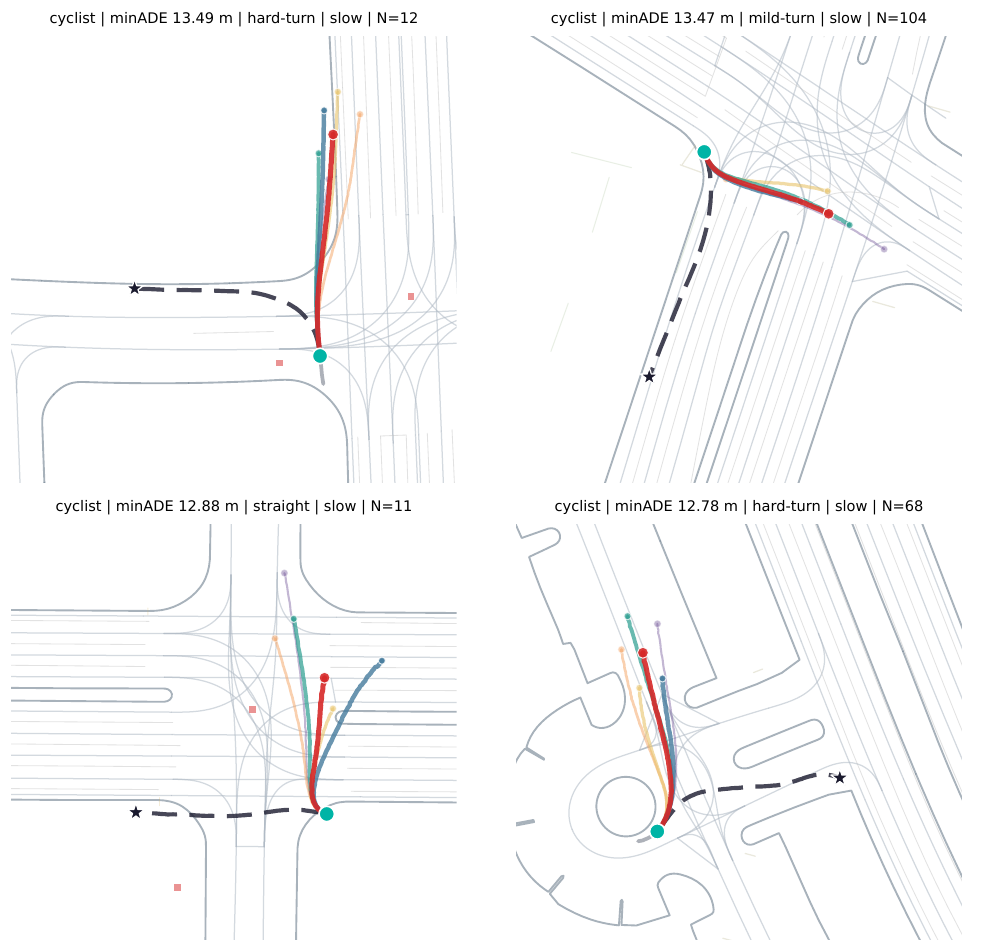}
    \caption{\textbf{Worst cyclist failure cases} (ranked by minADE). GT (dashed), history, and the top-3 predicted modes over the HD map. Failures are dominated by rare high-curvature intersection turns.}
    \label{fig:failure_cyclist}
\end{figure}

\section{Limitations}
The learned prior emits $K\!=\!64$ anchor trajectories per agent (one per decoder query, $N_q\!=\!K$), which greedy NMS prunes to the $6$ hypotheses required by the official WOMD protocol; the binding limitation is therefore the coverage of tail-distribution maneuvers (e.g., rare cyclist U-turns) within the retained set rather than a literal six-component mixture. This is most visible for cyclist maneuvers (Fig.~\ref{fig:failure_cyclist}): cyclists have the lowest per-class mAP, as the retained hypotheses under-represent rare cyclist behaviors such as sidewalk-to-road transitions. Scaling the anchor count $K$ improves coverage but increases memory and latency, a trade-off quantified in the anchor-count ablation (Tab.~\ref{tab:k_ablation}). Our evaluation is limited to WOMD; transfer to datasets with different map representations, shorter horizons, or different agent densities would require encoder adaptation and prior recalibration. The current formulation predicts marginal per-agent futures without explicit joint consistency, which may produce overlapping trajectories in dense interactions (though overlap rate remains competitive; Tab.~\ref{tab:main_test}). On the WOMD \emph{interactive} (joint) split, FlowS attains a joint single-step mAP of $0.2230$ (minADE $0.975$); as expected for a marginal predictor this trails joint-trained methods, since per-agent accuracy transfers only partially to scene-consistent joint hypothesis sets, which motivates the extension below. Extending FlowS to joint prediction via interaction-conditioned flows is a natural next step.

\section{Conclusion}
We have identified a conditioning strategy for one-step generative prediction: \emph{single-step integration is reliable when the transport problem is conditioned to be local.} FlowS implements this with a scene-conditioned prior that converts mode discovery into local refinement, and a semigroup-consistent displacement field that ensures one-step accuracy. Anchoring the displacement field at learned priors along straight-line paths yields a stable, low-variance self-consistency target, reducing the training instability of prior formulations on curved diffusion paths.

On the Waymo Open Motion Dataset, FlowS reduces iterative refinement to a single deterministic forward pass at $\sim$75\,FPS while achieving state-of-the-art mAP. Future work will extend the local transport conditioning to joint multi-agent flows and closed-loop simulation. An additional direction is to learn a reverse mapping from data back to the anchor space and enforce bijectivity between the forward and reverse transport steps, which could further regularize the displacement field.


\appendices
\section{Prior Head Implementation and Hyperparameters}
\label{app:impl}
The learned prior parameterizes each mode with $K$ learned queries $\{q_k\}$ applied to a hidden scene representation $h(c)\!\in\!\mathbb{R}^{D'}$, obtained by encoding the per-agent context $c$ through a small residual network, via FiLM~\cite{perez2018film}: $z_k=\gamma_k\odot h(c)+\beta_k$. From $z_k$, lightweight heads predict the mean trajectory $\mu_k$ (with waypoints refined by a small 1D temporal decoder) and the per-timestep scales $\sigma_k$. This realizes the mode-specialization step summarized in Sec.~\ref{subsec:learned_prior}, Eq.~\eqref{eq:prior_anchor}.

\emph{Default hyperparameters:} $D\!=\!256$-dim context; a $2$-layer residual prior MLP with $192$ hidden units and dropout $0.1$; per-timestep $\sigma$ with a log-$\sigma$ ramp from $-1.5$ (near) to $-0.5$ (far) and sampling range $\sigma\!\in\![0.3,0.8]$; anchor sampling temperature $\tau\!=\!0.5$; hinge diversity margin $m\!\in\![3,6]$\,m.

\bibliographystyle{IEEEtran}
\bibliography{IEEEabrv,egbib_compact}

\end{document}